\documentclass[11pt,conference]{IEEEtran}
\IEEEoverridecommandlockouts

\usepackage{cite}
\usepackage{amsmath,amssymb,amsfonts}
\usepackage{algorithmic}
\usepackage{graphicx}
\usepackage{textcomp}
\usepackage{xcolor}

\usepackage[hidelinks]{hyperref}
\usepackage{xurl}
\def\BibTeX{{\rm B\kern-.05em{\sc i\kern-.025em b}\kern-.08em
    T\kern-.1667em\lower.7ex\hbox{E}\kern-.125emX}}
\begin{document}

\title{DeepGI: Explainable Deep Learning for Gastrointestinal Image Classification\\
%{\footnotesize \textsuperscript{*}Note: Sub-titles are not captured for https://ieeexplore.ieee.org  and
}

\author{
    Walid Houmaidi, Mohamed Hadadi, Youssef Sabiri, Yousra Chtouki \\
    School of Science and Engineering, Al Akhawayn University \\
    Ifrane, Morocco \\
    Email: \{w.houmaidi, m.hadadi, y.sabiri, y.chtouki\}@aui.ma
}

\maketitle

\begin{abstract}
This paper presents a comprehensive comparative model analysis on a novel gastrointestinal medical imaging dataset, comprised of 4,000 endoscopic images spanning four critical disease classes: Diverticulosis, Neoplasm, Peritonitis, and Ureters. Leveraging state-of-the-art deep learning techniques, the study confronts common endoscopic challenges such as variable lighting, fluctuating camera angles, and frequent imaging artifacts. The best performing models, VGG16 and MobileNetV2, each achieved a test accuracy of 96.5\%, while Xception reached 94.24\%, establishing robust benchmarks and baselines for automated disease classification. In addition to strong classification performance, the approach includes explainable AI via Grad-CAM visualization, enabling identification of image regions most influential to model predictions and enhancing clinical interpretability. Experimental results demonstrate the potential for robust, accurate, and interpretable medical image analysis even in complex real-world conditions. This work contributes original benchmarks, comparative insights, and visual explanations, advancing the landscape of gastrointestinal computer-aided diagnosis and underscoring the importance of diverse, clinically relevant datasets and model explainability in medical AI research.
\end{abstract}


\section{Introduction}

Gastrointestinal (GI) diseases represent a major health burden globally, with over 2.3 billion cases and 2.56 million annual deaths attributed to digestive disorders \cite{frontiers_burden2023}; in Europe alone, GI and liver disorders account for nearly one million deaths each year \cite{ueg_survey2025}. Conditions such as Diverticulosis, Neoplasm, Peritonitis, and Ureters can result in severe complications, require urgent intervention, and are often difficult to diagnose early due to their variable presentation and anatomical location.

Historically, conventional endoscopy has been the gold standard for diagnosing GI diseases. While effective, it is invasive, requires sedation, and often limits accessibility for patients in rural or resource-limited settings. Operator variability, incomplete visualization of the mid-small bowel, and inability to perform therapeutic interventions during capsule endoscopy further compound the challenge \cite{pmc_nih_capsule2016}. Diagnostic yield is also affected by the quality of images, missed lesions due to rapid instrument movement, and psychological barriers faced by patients. As a result, clinicians must frequently rely on time-consuming manual review, which is prone to human error and requires extensive expertise.

Recent advances in artificial intelligence (AI), particularly deep learning and computer vision techniques, have revolutionized medical image analysis across domains such as neurology, hematology, and ophthalmology. AI-powered diagnostic tools have dramatically accelerated image interpretation. For example, VUNO Med®-BoneAge measures bone age from X-rays in five seconds, and Aidoc's triage algorithms have reduced turnaround times in emergency radiology by up to 30\%, matching or outperforming human experts in accuracy for various imaging tasks \cite{lancet_dl_vs_human2019, sciencedirect_ai_diagnostic2025}. These methods are now being applied to GI imaging, showing promise in both speed and precision.

However, existing public GI datasets and diagnostic models primarily focus on common anatomical features and exclude rarer, clinically significant conditions. To address this gap, we employ the Medical Imaging Dataset, which, to the best of our knowledge, has not been previously studied or benchmarked in the literature. The dataset comprises 4,000 endoscopic images deliberately curated to represent Diverticulosis, Neoplasm, Peritonitis, and Ureters disease classes that pose unique diagnostic challenges and demand robust computational approaches.

In this study, we train and evaluate three state-of-the-art CNN architectures using this focused dataset, establishing the first performance benchmarks and analyses for these diseases. We also incorporate explainable AI methodologies, notably Gradient weighted Class Activation Mapping (Grad-CAM), to interpret model predictions visually and facilitate clinical trust. Our contributions advance AI-driven GI disease diagnosis, enrich publicly available benchmarks, and highlight the urgent need for diverse datasets and interpretable models in medical imaging.

The remainder of this paper is organized as follows: Related Work, Methodology, Results And Discussion, Explainable AI Findings, and Conclusion And Future Work.

\section{Related Work}

Significant progress in computer vision and deep learning has transformed medical image analysis across multiple disciplines, notably in cerebral diagnostics \cite{houmaidi2025brainfusion}, hematological annotation \cite{atrrigen2025}, and ophthalmic disease detection \cite{eyedex2025}. Within gastrointestinal (GI) medicine, endoscopic imaging remains the gold standard for diagnosis, surveillance, and therapeutic intervention, posing substantial challenges due to variations in lighting, perspectives, patient anatomy, and frequent imaging artifacts \cite{BMJEndoscopy2025, PubMedDatasets2023}.

The last decade has seen a proliferation of public GI endoscopy datasets designed to support the research community in developing reliable AI models and benchmarks. Reviews enumerating 40+ endoscopic datasets highlight disproportionate focus on polyps and anatomical landmarks, with colonoscopy and capsule endoscopy dominating the offerings \cite{PubMedDatasets2023}. Kvasir and HyperKvasir are among the most widely adopted, containing tens of thousands of annotated images for tasks such as polyp, ulcer, and anatomical classification \cite{Pogorelov2017Kvasir, HyperKvasir2021}.

Despite advances, substantial gaps remain, particularly for rare or diagnostically challenging pathologies such as diverticulosis and peritonitis, which are notably absent or underrepresented in leading datasets \cite{NIHDatasetsReadiness2025}. Recent works argue the necessity for dataset diversity in disease representation, equipment type, and demographics to support robust, generalizable AI models \cite{NIHDatasetsReadiness2025}.

Studies leveraging CNNs for GI endoscopic image classification and segmentation have yielded impressive performance. Multiple architectures such as AlexNet, DenseNet, ResNet, and GoogLeNet, EfficientNet, and Vision Transformers are benchmarked with robust results, often exceeding 90\% accuracy on multiclass problems \cite{PeerJCNN2023, SciDirectComparative2024, JISTCNNViT2024, FrontiersGastric2024}. Ensemble and hybrid architectures combining CNNs with FFNNs and gradient-based segmentation have further improved diagnostic accuracy, reaching F1-scores and accuracies above 97\% \cite{PeerJCNN2023, SciDirectComparative2024}.

However, attention is increasingly paid to external validity; most models excel in heavily benchmarked categories but lag in less common clinical scenarios due to data imbalance and limited disease representation \cite{BMJEndoscopy2025}. Innovations such as EndoDINO, which leverages pre-training on vast diverse datasets, indicate foundation modeling and self-supervised learning may overcome some generalizability barriers \cite{EndoDINO2025}.

For rare classes, optimized CNNs and Vision Transformers have demonstrated potential in structured comparative studies. Ayan (2024) showed DenseNet201 outperforming ViT with optimized transfer learning, achieving 93.13\% accuracy, recall, and F1-score \cite{JISTCNNViT2024}. Rahman et al. (2024) presented EfficientNetB5 strategies with sophisticated augmentation, highlighting robustness to real-world imaging variability \cite{PMCEnhanceImg2024}. Ensemble methods are also adopted for more consistent clinical performance in practical GI diagnosis \cite{PeerJEnsembleGIT2025}.

In summary, while deep learning has revolutionized GI image analysis, limitations related to dataset composition, disease diversity, and external validity persist. The medical imaging dataset investigated in this work, focused on four key underrepresented GI classes, directly addresses these research gaps and is positioned as a vital resource for advancing computer vision and clinical diagnostic methodologies in GI endoscopy.

\section{Methodology}

\subsection{Proposed System Architecture}
The proposed system leverages deep convolutional neural networks to classify gastrointestinal endoscopy images into four distinct categories: Diverticulosis, Neoplasm, Peritonitis, and Ureters. Inference begins with a raw endoscopic image input, which is automatically standardized and preprocessed. The image is then processed through a selected trained model (VGG16, MobileNetV2, or Xception), which extracts robust features and predicts the disease class in real time. Automated augmentation pipelines and input normalization help maintain model consistency and accuracy under challenging imaging conditions. The entire workflow is optimized for rapid inference, enabling clinicians to obtain immediate diagnostic predictions and visual explanations for each case, thus supporting practical integration in clinical decision making.\cite{pmc6775529,pmc11601128,pmc10572467}.

\subsection{Dataset Description}
\begin{figure}[ht]
  \centering
  \includegraphics[width=0.5\textwidth]{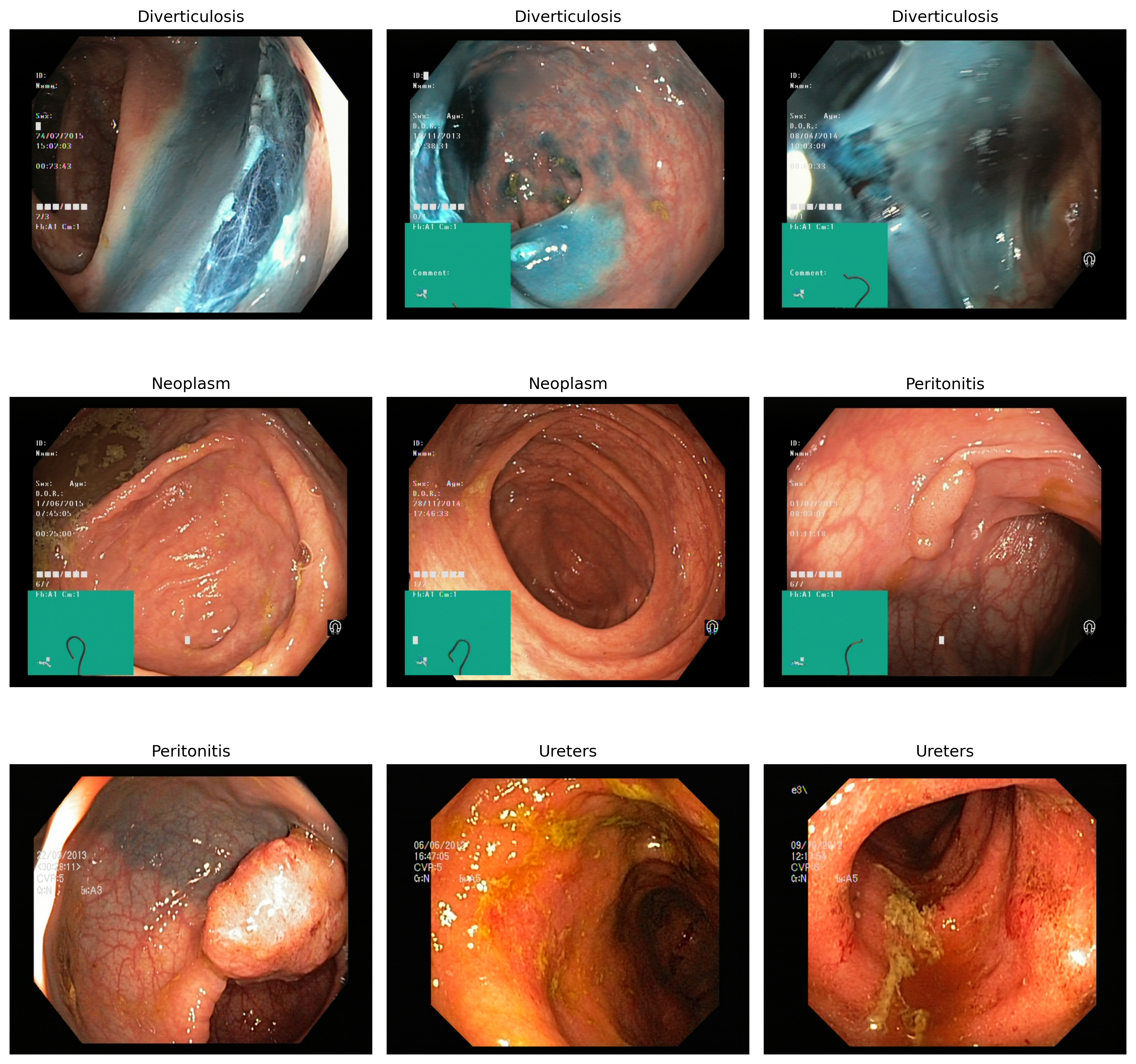}
  \caption{Representative endoscopic images from the four disease classes: Diverticulosis, Neoplasm, Peritonitis, and Ureters.}
  \label{fig:sampledata}
\end{figure}

The dataset uniquely emphasizes conditions that are underrepresented in existing public GI imaging benchmarks, offering an invaluable resource for the development and validation of advanced diagnostic algorithms. The dataset used in this study is publicly available on Kaggle~\cite{heartzhacker2023medicalimaging} and contains high-resolution gastrointestinal endoscopic images. It is organized into four classes relevant to rare but clinically significant GI diseases: Diverticulosis, Neoplasm, Peritonitis, and Ureters. Figure \ref{fig:sampledata} illustrates sample labeled images from the datatset. Each class consists of approximately 1,000 images curated from real-world procedures, presenting challenges such as lighting variations and imaging artifacts. Images are resized and normalized to a fixed input size of $224 \times 224 \times 3$ (RGB channels) before model ingestion.

\subsection{Data Augmentation}
To improve generalization and robustness, extensive data augmentation is employed, simulating variations encountered in endoscopic imaging, these include:
\begin{itemize}
    \item \textbf{Zoom range:} 0.3
    \item \textbf{Shear range:} 0.4
    \item \textbf{Rotation range:} $20^\circ$
    \item \textbf{Width shift:} 0.2
    \item \textbf{Height shift:} 0.2
    \item \textbf{Brightness range:} [0.6, 1.4]
    \item \textbf{Horizontal flip:} True
    \item \textbf{Fill mode:} Nearest
    \item All images converted to $224 \times 224$ RGB.
\end{itemize}
This approach helps the models handle diverse real-world imaging scenarios, including artifacts and scanner differences \cite{pmc7778711, arxiv2402}.

\subsection{Model Selection and Justification}
\textbf{VGG16:} Widely adopted for medical imaging due to its deep architecture and proven feature extraction capabilities, VGG16 excels in classification tasks, especially in datasets with moderate scale and complexity, yielding robust and interpretable results \cite{pmc6775529}.

\textbf{MobileNetV2:} Chosen for its lightweight structure and high classification accuracy, MobileNetV2 is well suited for deployment in resource-constrained environments, such as real-time clinical decision support systems \cite{pmc10572467}.

\textbf{Xception:} Utilizes depthwise separable convolutions, enhancing feature extraction while reducing computational cost. Xception achieves remarkable performance on fine-grained medical image classification tasks and handles variability in imaging modalities effectively \cite{pmc11601128}.

\subsection{Evaluation Metrics}
Each model is compared using overall accuracy and confusion matrix analysis. Accuracy quantifies the proportion of correct predictions in the test set. Confusion matrices enable granular evaluation of model performance per class, identifying strengths and weaknesses in disease discrimination \cite{sciencedirect151}.

\subsection{Model Architectures and Fine-Tuning}
The following training and fine-tuning steps were applied to each neural network model in this study:

\begin{itemize}
    \item All base architectures (MobileNetV2, VGG16, Xception) were initialized with weights pre-trained on ImageNet.
    \item Input images were resized to $224 \times 224$ pixels and normalized to 3 channels (RGB).
    \item Custom classification layers were appended, including dense layers with ReLU activation, batch normalization, dropout (rate $=0.5$), and L2 regularization ($\lambda=0.001$).
    \item The models were compiled using the Adam optimizer with a learning rate of $1\times10^{-5}$, categorical cross-entropy loss, and accuracy as the evaluation metric.
    \item Training was conducted for 100 epochs, with batch size selected based on available computational resources.
    \item A learning rate scheduler (ReduceLROnPlateau) reduced learning rate by a factor of 0.5 if validation loss did not improve after 3 epochs (minimum rate $=1\times10^{-6}$).
    \item Early stopping was monitored on validation loss; training was halted after 7 stagnant epochs, restoring the best-performing weights.
    \item Model checkpointing was used to save the architecture and weights only when validation accuracy improved.
    \item All base layers were frozen for initial training, then selectively unfrozen for the final few layers to enable domain-specific fine-tuning.
    \item The last layer utilized \texttt{softmax} activation to output class probabilities for gastrointestinal disease classification.
\end{itemize}

%findings and discussion section
\section{Results and Discussion}

\subsection{Performance Comparison}
The classification performance of the three convolutional neural network models (MobileNetV2, VGG16, and Xception) was evaluated on the gastrointestinal dataset as stated in Table 1. MobileNetV2 and VGG16 both achieved a test accuracy of 96.5\%, while Xception reached a slightly lower test accuracy of 94.24\%. This result indicates that while all models demonstrate strong performance, Xception’s architecture may not be as effective as MobileNetV2 and VGG16 for capturing the relevant visual features in this dataset, highlighting the importance of model selection for disease classification tasks.

\begin{table}[h]
\centering
\caption{Test Accuracy Comparison of CNN Models}
\resizebox{0.7\linewidth}{!}{%
\begin{tabular}{|c|c|}
\hline
\textbf{Model} & \textbf{Test Accuracy (\%)} \\
\hline
MobileNetV2 & \textbf{96.5} \\
VGG16 & \textbf{96.5} \\
Xception & 94.24 \\
\hline
\end{tabular}%
}
\end{table}

\subsection{Results Interpretation}
All three models demonstrated strong classification performance, indicating that CNNs are well suited for rare gastrointestinal disease detection from endoscopic images. MobileNetV2 and VGG16 exhibited almost identical results at the top accuracy tier, which may be attributed to their efficient feature extraction and powerful transfer learning capabilities. In contrast, Xception reached a slightly lower test accuracy of 94.24\%. This outcome suggests that while Xception’s depthwise separable convolutions offer advanced feature representation, they may not always generalize better for this specific dataset or task. These findings underscore that lightweight models such as MobileNetV2 can compete favorably even outperform deeper architectures on moderately sized medical image datasets, making them attractive for both performance and computational efficiency.

\subsection{Classification Report of VGG16}
\begin{table}[h!]
\centering
\caption{Classification Report for VGG16 Model}
\begin{tabular}{|l|c|c|c|c|}
\hline
\textbf{Class}        & \textbf{Precision} & \textbf{Recall} & \textbf{F1-score} & \textbf{Support} \\
\hline
Diverticulosis & 0.99 & 0.99 & 0.99 & 200 \\
Neoplasm       & 0.95 & 0.97 & 0.96 & 200 \\
Peritonitis    & 0.96 & 0.94 & 0.95 & 200 \\
Ureters        & 0.95 & 0.96 & 0.96 & 200 \\
\hline
\textbf{Accuracy}     &      &      & 0.965 & 800 \\
\textbf{Macro avg}    & 0.96 & 0.96 & 0.96 & 800 \\
\textbf{Weighted avg} & 0.96 & 0.96 & 0.96 & 800 \\
\hline
\end{tabular}
\end{table}

The VGG16 model achieved highly balanced and robust multi-class classification performance across all four gastrointestinal disease categories. Table 2 illustrates detailed classification report of VGG16 on test set. Precision values ranged from 0.95 to 0.99, reflecting the model’s strong ability to minimize false positives, while recall scores between 0.94 and 0.99 show effective detection of true disease cases. Diverticulosis performed slightly better across all metrics, likely due to distinctive imaging features, but Neoplasm, Peritonitis, and Ureters also exhibited high precision and recall, confirming the model’s resilience in classifying both common and challenging classes. The overall validation accuracy of 96.5\% and consistent macro/weighted averages indicate that VGG16 delivers reliable, unbiased predictions and leverages deep feature extraction for generalized clinical endoscopic image analysis.

\begin{figure}[ht]
    \centering
    \includegraphics[width=0.5\textwidth]{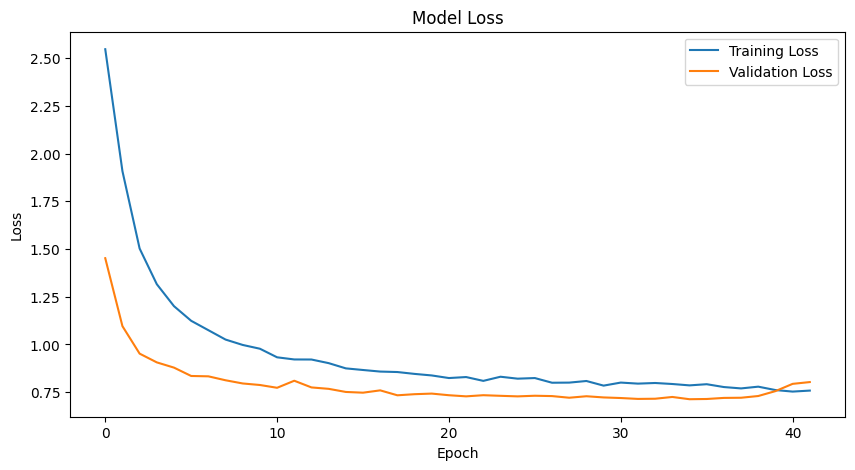}
    \caption{Training and validation loss curves for the VGG16 model over 42 epochs. The plot demonstrates steady convergence and generalization performance during training, with early stopping triggered at epoch 42.}
    \label{fig:vgg16_loss}
\end{figure}
\noindent
The training and validation loss curves for the VGG16 model (Figure~\ref{fig:vgg16_loss}) illustrate steady convergence throughout training. Both losses decrease rapidly during early epochs and then plateau, suggesting effective optimization. The small gap between training and validation loss indicates successful generalization and minimal overfitting.

\begin{figure}[ht]
    \centering
    \includegraphics[width=0.5\textwidth]{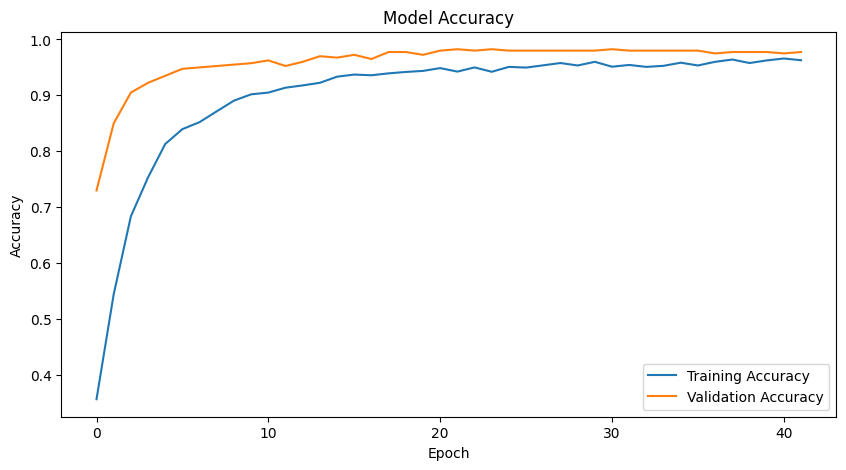}
    \caption{Training and validation accuracy curves for the VGG16 model. The validation accuracy closely follows the training accuracy, indicating stable and effective learning throughout the epochs,with early stopping triggered at epoch 42.}
    \label{fig:vgg16_acc}
\end{figure}
\noindent
Figure~\ref{fig:vgg16_acc} presents the training and validation accuracy curves for the VGG16 model. The validation accuracy closely follows the training accuracy, demonstrating the model's strong generalization and effective learning. The high validation accuracy achieved across epochs underscores the model's capability to accurately classify gastrointestinal images.

\section{Explainable AI Findings}

\begin{figure}[ht]
    \centering
    \includegraphics[width=0.5\textwidth]{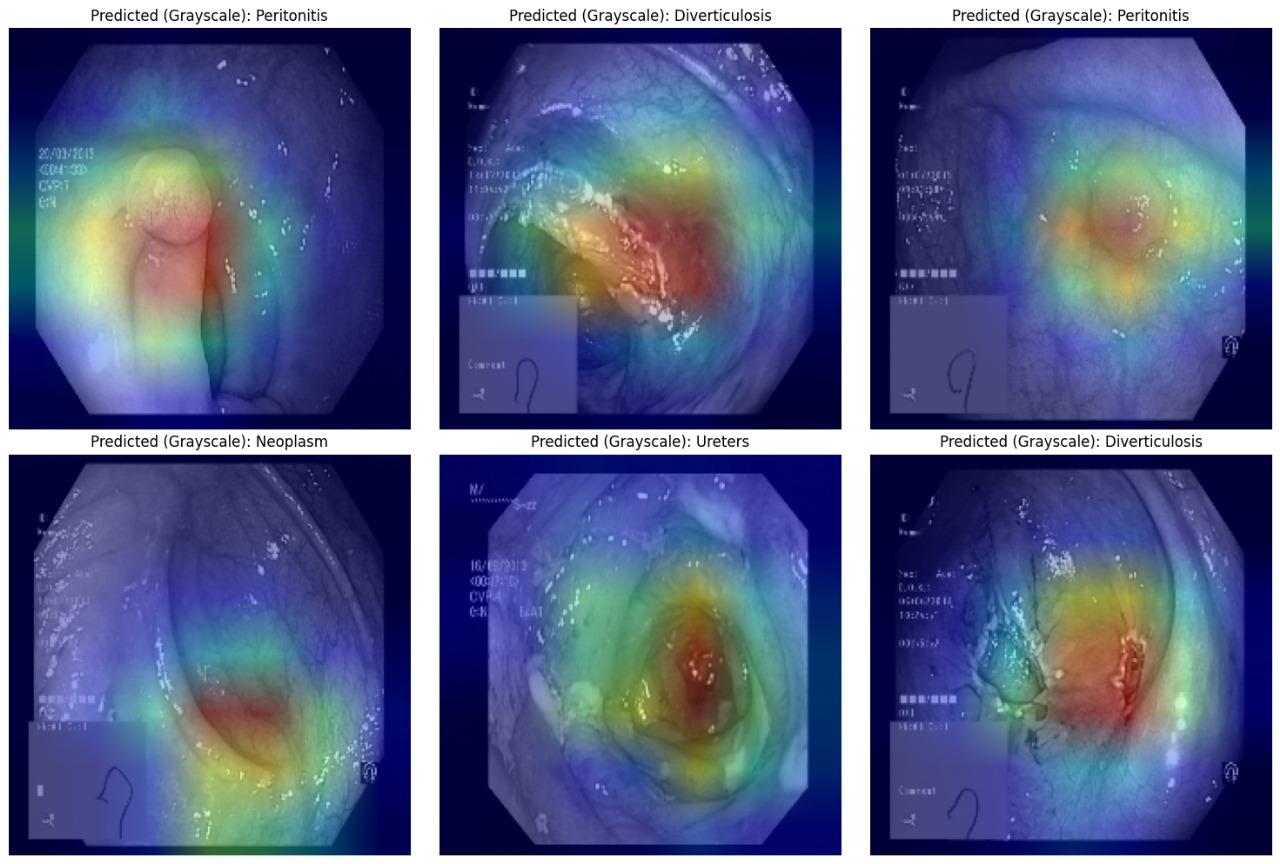}
    \caption{Grad-CAM heatmap overlays for selected endoscopic images. The highlighted regions correspond to areas most influential in the automated prediction for each disease class.}
    \label{fig:explainable_ai}
\end{figure}

To enhance the interpretability of disease predictions and increase clinical trust in automated analysis, the proposed system incorporates explainable AI (XAI) visualization techniques. Figure~\ref{fig:explainable_ai} presents representative examples of heatmap overlays generated by Grad-CAM, illustrating the regions of each endoscopic image that most strongly influence the model's classification across four disease classes: Peritonitis, Diverticulosis, Neoplasm, and Ureters.

These visual explanations were generated using the Gradient-weighted Class Activation Mapping (Grad-CAM) technique, which highlights the most influential regions in each endoscopic image for the network's predictions. To further enhance visual clarity and interpretability, the original endoscopic images were converted to grayscale before overlaying the Grad-CAM heatmaps. This pre-processing step reduces color distractions and allows the highlighted decision areas to stand out more distinctly, aiding both quantitative assessment and qualitative evaluation by medical practitioners.

Notably, the heatmaps consistently focus on clinically relevant regions, confirming that the model leverages meaningful visual cues rather than unrelated artifacts. This supports the system's reliability and potential integration into real-world diagnostic workflows, providing actionable feedback for medical practitioners.

Such XAI visualizations not only bolster transparency, but also facilitate the identification of possible model biases or errors, promoting an iterative approach to model refinement and enhancing user confidence in automated gastrointestinal disease diagnosis.

\section{Conclusion and Future Work}

This study presented a comparative analysis of deep convolutional neural network models (MobileNetV2, VGG16, and Xception) for the classification of gastrointestinal disease images. Experimental results demonstrated that VGG16 and MobileNetV2 achieved the highest test accuracies, while Xception performed marginally lower, underscoring the importance of model selection and fine-tuning in medical image analysis. Key training strategies such as early stopping, learning rate scheduling, and regularization contributed significantly to model robustness and generalization.

Despite promising outcomes, some challenges remain, particularly regarding the handling of diverse clinical imaging artifacts and dataset limitations. Future work will focus on expanding the dataset, exploring advanced architectures and data augmentation techniques, and integrating cross-validation frameworks for further performance enhancement. Developing interpretable AI models and investigating transfer learning across related medical imaging tasks are also prioritized directions, aiming to support real-world deployment in clinical environments.

In summary, this work advances automated gastrointestinal disease detection and offers a foundation for future research in intelligent healthcare image analysis.

\bibliographystyle{IEEEtran}
\bibliography{references}

\end{document}